\pdfoutput=1
\documentclass{article}
\usepackage[final]{colm2026_conference}

\usepackage{wrapfig}
\usepackage{microtype}
\usepackage{hyperref}
\usepackage{url}
\usepackage{booktabs}
\usepackage{multirow}
\usepackage{graphicx}
\usepackage{listings}
\usepackage{subcaption}
\usepackage[most]{tcolorbox}

\lstdefinestyle{prompt}{
    basicstyle=\ttfamily\footnotesize,
    breaklines=true,
    breakatwhitespace=true,
    columns=fullflexible,
    frame=single,
    framerule=0.8pt,
    framextopmargin=2pt,
    framexbottommargin=2pt,
    xleftmargin=0pt,
    xrightmargin=0pt,
    captionpos=b,
    extendedchars=true,
    literate=%
        {—}{{---}}1
        {–}{{--}}1
        {“}{{``}}1
        {”}{{''}}1
        {‘}{{`}}1
        {’}{{'}}1
        {"}{{\char34}}1
        {×}{{$\times$}}1
        {μ}{{$\mu$}}1
        {Erdős}{{Erd\H{o}s}}1
}

\usepackage{lineno}

\definecolor{darkblue}{rgb}{0, 0, 0.5}
\hypersetup{colorlinks=true, citecolor=darkblue, linkcolor=darkblue, urlcolor=darkblue}
\newcommand{\rateci}[3]{\shortstack{#1\%\\[-2pt]{\scriptsize [#2, #3]}}}

\definecolor{cardfill}{gray}{0.985}
\definecolor{panelfill}{gray}{0.93}
\definecolor{authorfill}{rgb}{0.90,0.92,0.97}

\tcbset{
  exhibitbase/.style={
    enhanced, breakable,
    colback=cardfill, colframe=darkblue, boxrule=0.6pt,
    arc=2pt, outer arc=2pt,
    left=8pt, right=8pt, top=6pt, bottom=6pt,
    fonttitle=\sffamily\bfseries\small, coltitle=white, colbacktitle=darkblue,
  },
}

\newcommand{\qsec}[1]{\par\smallskip{\sffamily\bfseries\footnotesize\color{darkblue}#1}\par\nobreak\vspace{2pt}}

\newcommand{\optionpanel}[3][panelfill]{%
  \setlength{\fboxsep}{6pt}%
  \noindent\colorbox{#1}{\parbox[t]{\dimexpr\linewidth-12pt\relax}{\small\textbf{#2}~#3}}%
}

\newtcolorbox{querycard}[2][]{exhibitbase, fontupper=\small, title={#2}, #1}

\tcbset{
  respinner/.style={enhanced, arc=2pt, colback=white, colframe=gray!45,
    boxrule=0.5pt, fonttitle=\sffamily\bfseries\footnotesize, coltitle=black, colbacktitle=panelfill,
    left=6pt, right=6pt, top=4pt, bottom=4pt},
  respauthor/.style={respinner, colbacktitle=authorfill, colframe=darkblue},
}

\newtcblisting{promptbox}[1][]{%
  exhibitbase, listing only,
  listing options={style=prompt, frame=none, framerule=0pt}, #1
}

\title{Eliciting Research Taste in LLMs through Future Research Direction Choice}

\author{Dhruv Trehan, Paras Chopra \\
Lossfunk\\
\texttt{\{dhruv.trehan, paras\}@lossfunk.com} \\
}

\begin{document}

\ifcolmsubmission
\linenumbers
\fi

\maketitle

\begin{abstract}
Research taste is increasingly discussed in the context of AI scientists, yet remains little studied or measured in LLMs. This judgment matters especially in open-ended research, where a direction must be chosen before a verifiable objective is available. We operationalize research taste through one task of the scientific workflow: choosing which future research direction to pursue given a research goal and current results. Rather than predicting downstream scientometric metrics or treating author choices as objectively correct, we evaluate models against human-authored future-work directions and report author alignment, stable divergence, and unstable choice as behavioral measures of research-direction preference. We construct an order-controlled evaluation of 80 pairwise comparisons from 19 ML papers published between December 2025 and February 2026 across four subdomains. Each query presents a research goal, hypotheses tested with key results, and two candidate future directions: one author-recommended direction and one plausible distractor sampled from a similar paper. Across eight LLMs, author alignment ranges from 45.0\% to 67.5\%, while stable divergence ranges from 15.0\% to 36.2\%. These divergences largely persist under similarity controls, prompt ablations, and an author-preference framing check, and inter-model agreement suggests that some non-author preferences are shared across models. We characterize what models diverge toward using transition-category labels and selected reasoning examples, finding recurring movement toward scaling/generalization and trade-offs between methodological and infrastructure/data extensions. We view this as an early descriptive evaluation for eliciting research taste profiles in LLMs, and as a step toward understanding and steering their research preferences as they take on more autonomous roles in novel open-ended scientific discovery. The evaluation set, all model responses, and analysis code are available at \url{https://github.com/Lossfunk/research-taste-eval-v1}.
\end{abstract}

\section{Introduction}

LLMs and LLM agents are making rapid progress on autonomous discovery workflows, but humans still choose the problems they target and shape their motivation. Recent agentic systems such as Autoresearch~\citep{karpathy2026autoresearch} and AlphaEvolve~\citep{novikov2025alphaevolvecodingagentscientific}, and their extensions~\citep{huggingface2026mlintern, kulikov2026autodata, sharma2025openevolve, zhang2025darwingodelmachine}, suggest that some forms of discovery can be automated through evolutionary search and closed-loop iteration over code and experimentation grounded in verifiable metrics. However, most scientific research is more open-ended, requiring an earlier choice on which research branches, limitations, or extensions to prioritize before a reliable short-horizon objective is available.

This upstream choice is closely related to what scientists often call research taste. There has been a surge in recent discussion about research taste and its lack in large language models, but limited work has systematically studied the concept or measured it behaviorally in humans and LLMs. Recent work on measuring and training research taste in LLMs focuses on predicting downstream scientometric metrics, such as citation counts and journal decisions, as proxies for impact. However, previous studies on the Science of Science have shown that these metrics capture author and lab visibility, venue prestige, and network effects more than research quality itself~\citep{doi:10.1126/science.aao0185}. These are also lagging metrics and materialize years downstream, making time-controlled evaluation design and avoiding training data contamination a real challenge for this work. 

In this work, we propose an alternative small task and pipeline for eliciting research taste signals in LLMs, namely selecting which future research direction to pursue, given an existing research context. Figure~\ref{fig:example_query} shows an example query. 

We make three contributions. First, we build a pairwise evaluation of 80 such queries that uses human-authored future works as ground-truth reference points matched with distinct and plausible distractors. This lets us study when models align with author-recommended directions, when they stably prefer non-author alternatives, and when their choices are sensitive to option ordering. Second, we evaluate eight frontier and open-weight models and show that non-author preferences, especially stable divergences, recur across models and queries and are not explained by simple similarity signals. Third, we study the structure of these divergences through prompt variations, inter-model agreement analysis, an early descriptive taxonomy of model research-direction preferences, and selected model responses.

\begin{figure}[h]
\begin{querycard}[breakable=false, fontupper=\footnotesize, left=6pt, right=6pt, top=4pt, bottom=4pt]{Example Query: Future-Work Selection}
\textit{\small You are an AI research assistant helping decide the next research direction for a running research project. Think step by step before answering.}

\qsec{Research Goal}
Develop an environment for training reinforcement learning-based search agents to answer scientific questions by searching and reasoning over biomedical research papers, including creating the necessary datasets, retrieval corpus, and benchmarks...

\qsec{Hypotheses Tested \& Key Results}

\textbf{H1.} RLVR training of search agents leads to stronger performance on scientific paper question-answering compared to non-RL baselines... \emph{For 3B models, RL improved over RAG by 9.6 and 5.5 points on the proposed dataset and a standard biomedical QA benchmark, and for 7B models the difference was 14.5 and 9.3 points.}\par\smallskip
\textbf{H2.} Factoid question-answer pairs with verifiable answers can serve as effective training data for RLVR methods for scientific literature search agents. \emph{The automatically generated dataset contained 54,907 training samples and enabled exact-match reward computation.}\par\smallskip
\textbf{H3.} Search agents trained with RLVR exhibit emergent behaviors including explicit planning about query rewriting, reasoning before retrieval, and verification when the model already has an initial answer. \emph{Manual review revealed explicit planning and keyword extraction, reasoning before search, and verification behavior.}\par\smallskip
\textbf{H4.} Paraphrasing questions during dataset construction to avoid exact keyword matching increases dataset difficulty and benefits robust search-agent training. \emph{For the RLVR agent, non-paraphrased questions scored 57.2\%, while paraphrased questions scored 44.9\%.}

\qsec{Which research direction should this project pursue next?}
\optionpanel[authorfill]{A.}{Equip agents with tools and metadata that real scientists use in their work, such as citation traversal capabilities and source reliability metrics.}\par\smallskip
\optionpanel{B.}{Incorporate human-in-the-loop capabilities that enable subject-matter experts to approve, revise, or redirect agent decisions to increase system reliability and alignment with real-world scientific practices.}
\end{querycard}
\caption{Example query (abridged). The model receives a research goal, tested hypotheses with key results, and two candidate next directions. Here Option A is author-recommended and Option B is the sampled distractor. Each pair is also evaluated in the reversed A/B ordering.}
\label{fig:example_query}
\end{figure}

\section{Evaluation Creation \& Setup}
\label{eval_creation_and_setup}

\textbf{Dataset Construction}. Our dataset of 80 pairwise comparisons (160 prompts) comes from 19 papers across 4 subdomains of ML research, namely Agents/Methods, LLM Analysis, Training/Infrastructure, and AI for Science. We started with a pool of 41 papers from December 2025-February 2026, after the stated training cutoff dates of the evaluated models, to mitigate data contamination. We prioritized papers from established research groups to ensure that the extracted research context (hypotheses, results, and future work) meets the standard methodological and rigor expectations of the human research community.

\begin{figure}[h]
\centering
\includegraphics[
    width=\linewidth,
    trim=20 30 30 15,
    clip
]{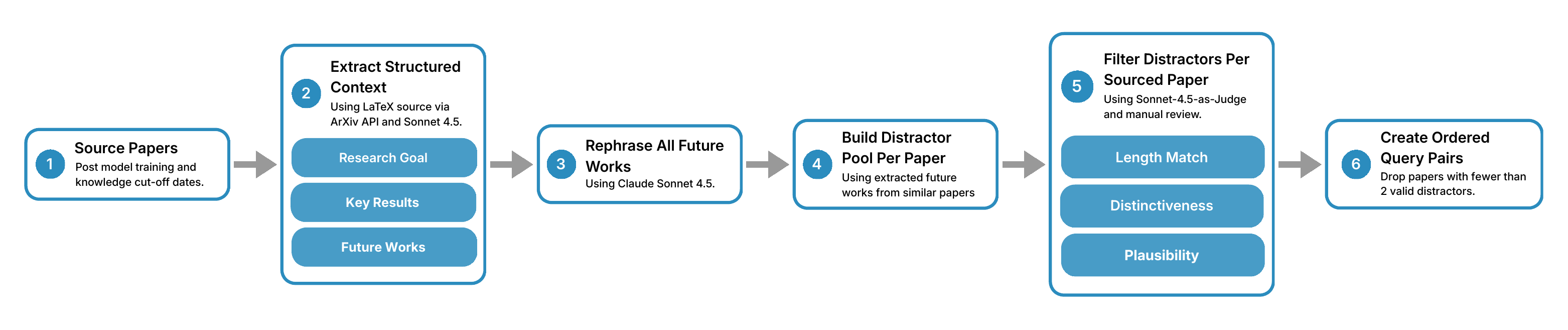}
\caption{Dataset construction and evaluation pipeline.}
\label{fig:evaluation_creation}
\end{figure}

Figure~\ref{fig:evaluation_creation} summarizes our dataset construction and evaluation pipeline, from source-paper selection through distractor filtering and ordered query-pair creation.

We access the LaTeX source for these 41 papers via the arXiv API and then filter them for explicitly stated future works using Appendix prompt~\ref{prompt:triage}. For each remaining paper, we use Claude Sonnet 4.5 at temperature 0.3 to extract three structured fields: the research goal, the hypotheses tested with key results, and future works with source citations. Prompts for each of these extractions are shared in Appendix~\ref{prompt:research_goal},~\ref{prompt:hypotheses}, and~\ref{prompt:future_works}. We also use Sonnet 4.5 to rephrase all extracted future works, using Appendix prompt~\ref{prompt:rephrasing}, into standardized research-direction statements while preserving original terminology where possible. This standardization reduces formatting and lexical cues that could otherwise distinguish author-recommended directions from distractors.

To construct a distractor pool for each target paper's ground-truth author-recommended future work statements, we compute an embedding similarity matrix across papers using OpenAI's text-embedding-3-small on an 8000-token excerpt of each paper's LaTeX source ( first 2{,}000 and last 6{,}000 tokens), and retrieve rephrased future works from the top-5 most similar papers as candidate distractors for each target paper. We filter candidates by length, requiring distractors to be within 0.75-1.25x the author-recommended future works word count, and use Claude Sonnet 4.5 as an LLM judge to assess distinctiveness from the target paper's ground-truth future works and plausibility as a next direction for the target paper (Appendix~\ref{prompt:distractor_eval}). After automated filtering and manual query-quality review, papers with fewer than 2 valid distractors are dropped. 19 of 41 papers are retained, yielding 80 unique pairwise comparisons.

We also piloted a three-option multiple-choice format before adopting this design. Pilot results are reported in Appendix Tables~\ref{tab:mcq_main_results} and~\ref{tab:mcq_position_bias}. While reviewing this, we ran into the limitations of multiple-choice evaluations and their susceptibility to position bias documented extensively in literature such as \citep{zheng2024largelanguagemodelsrobust}. To minimize position bias, our final query set was constructed pairwise.

As a dataset diagnostic, we compute embedding and TF-IDF similarity between each option and the research context. The author-recommended option is more context-similar in 64 of 80 pairs (80.0\%), but absolute similarities are low and the average delta is small relative to its variance, indicating that similarity alone is not a reliable discriminator. Full similarity statistics are reported in Appendix Table~\ref{tab:similarity_controls}.

\textbf{Evaluation Setup}. We evaluate Claude Opus 4.5 and Claude Haiku 4.5 from Anthropic, Gemini 3 Pro and Gemini 3 Flash from Google DeepMind, GPT-5.2 Pro and GPT OSS 120B from OpenAI, Qwen3 Max (Thinking) from Qwen, and Llama 4 Maverick from Meta. Models are chosen to span open and closed model providers and model size families. All evaluations use temperature 0 with one response per ordered prompt. The reported outcomes, therefore, represent single-sample behavioral elicitation. We leave multi-sample consistency and across-temperature checks, as well as log-probability-based confidence analysis and mechanistic interpretability analyses of these behaviors, to future work. 

We report three key metrics across 80 query pairs: author alignment (the author-recommended direction selected in both orderings), stable divergence (the sampled distractor selected in both orderings), and unstable choice (chosen option differs across the two orderings). We compute these metrics using only the final selected option and do not evaluate the soundness of the accompanying reasoning trace.

\section{Results}
\label{sec:results}

\subsection{Do models disagree with human authors on future research direction choices?}
\label{sec:headline_results}

\begin{figure}[h]
\centering
\includegraphics[width=0.80\linewidth]{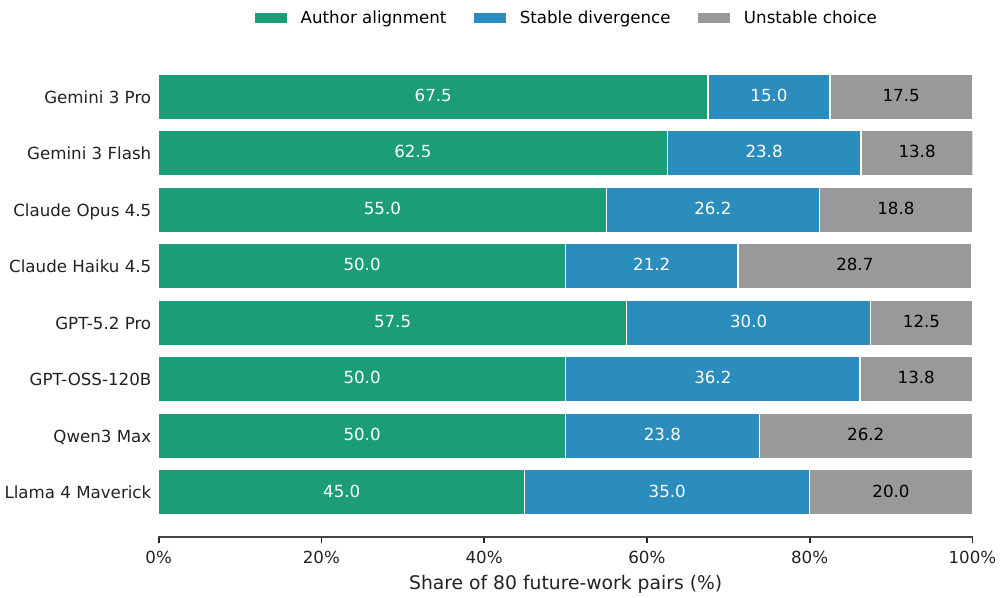}
\caption{Outcome shares for future-work selection across evaluated models. All models exceed the 25\% random baseline on author alignment and show some substantial stable divergence. Author alignment ranges from 45.0\% to 67.5\%, and stable divergence from 15.0\% to 36.2\%. Unstable-choice rates vary from 12.5\% to 28.7\%, indicating that some models preserve content choices across orderings more consistently than others.}
\label{fig:main_results_outcome_shares}
\end{figure}

Figure~\ref{fig:main_results_outcome_shares} summarizes the three outcome shares per model: Author Alignment, Stable Divergence, and Unstable Choice. Exact per-model values are reported in Appendix Table~\ref{tab:pairwise_results}. Among cases where models do not align with the author-recommended direction, models differ in whether they make an order-consistent content-level non-author choice or become sensitive to option ordering, producing unstable choices.

Stable divergence was not isolated to a select few queries. 45 of 80 pairs elicited stable divergence from at least one model, 34 from at least two, 27 from at least three, and 24 from at least four. Appendix table~\ref{tab:divergence_prevalence} reports the cumulative number of pairs eliciting stable divergence at each model-count. In total, we observe 169 stable divergence events across all models. These counts suggest that non-author preferences recur across models and queries, rather than remaining idiosyncratic to individual models.

\subsection{How often do models agree with each other?}
\label{intermodel_agreement}

We compute intermodel agreement across all 28 model pairs. Two models agree when both are in the same outcome state - both author-aligned, both stably divergent, or both unstable. Figure~\ref{fig:intermodel_three_state_agreement_heatmap} visualizes this.

Agreement in all three pairs of models from the same provider exceeds the mean of 61.9\%. However, the source of this agreement differs across providers. The OpenAI pair has a shared stable-divergence rate of 26.2\%, above both the across-pair mean of 15.6\% and the other within-provider pairs. By contrast, the Gemini and Claude within-provider pairs have lower shared-divergence rates of 11.2\% and 12.5\%, respectively, suggesting that their within-provider agreement is driven more by shared author alignment or shared unstable choices than by overlap in the same non-author choices. Complete pairwise agreement results, along with the shared divergence and shared unstable choice splits, are reported in Appendix Table~\ref{tab:intermodel_agreement}. 

\subsection{Are stable divergences just evaluation artifacts?}
\label{sec:divergence_ablations}

\textbf{Similarity control.} As discussed in Section~\ref{eval_creation_and_setup} Table~\ref{tab:similarity_controls} the author-recommended option is more context-similar in 64 of 80 pairs. This suggests a pure similarity heuristic would score an 80.0\% author-alignment rate. No evaluated model reaches this. To further verify that stable divergences reflect content-level preferences rather than similarity heuristics, we split per-model stable divergence events by their semantic similarity to author-recommended directions and extracted-distractors in Appendix table~\ref{tab:divergence_overlap}. Across all 169 stable-divergence events, \textbf{76.9\%} occur on pairs where the author-recommended option had higher similarity to context, with per-model rates ranging from 67\% to 82\%. 

\textbf{Prompt variation.} We re-ran response generation on all 169 stable divergence events with four prompt variations to explore the effect of prompting on these stable divergence events. Three of these variations were single-variable prompt ablations that modify exactly one element of the canonical query while holding all others identical - no `AI research assistant' role, no high-level `Research Goal', and a different task statement. The fourth was an alternative task framing that replaced the role and task to predicting which direction the researchers behind the original work would be more likely to choose for their own project. These prompts are included in Appendix~\ref{prompt:variations}.

\begin{figure}[h]
\centering
\begin{subfigure}[t]{0.48\linewidth}
    \centering
    \includegraphics[width=\linewidth]{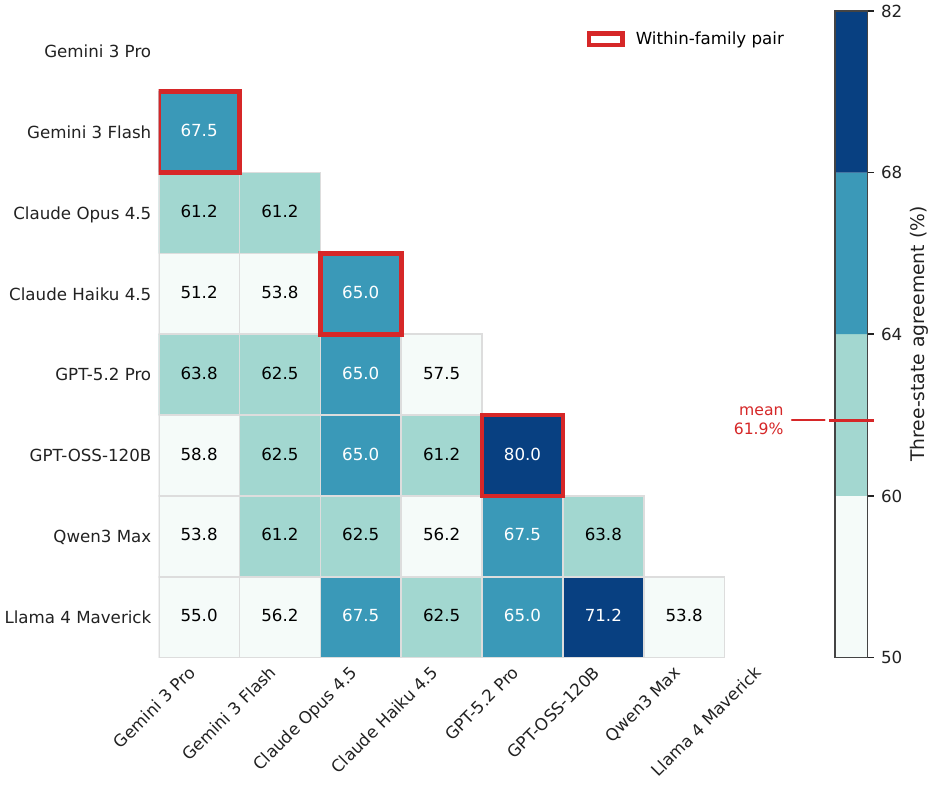}
    \caption{Inter-model agreement across three-state pairwise outcomes. Across all 28 model pairs, agreement ranges from 51.2\% to 80.0\% (mean 61.9\%).}
    \label{fig:intermodel_three_state_agreement_heatmap}
\end{subfigure}
\hfill
\begin{subfigure}[t]{0.48\linewidth}
    \centering
    \includegraphics[width=\linewidth]{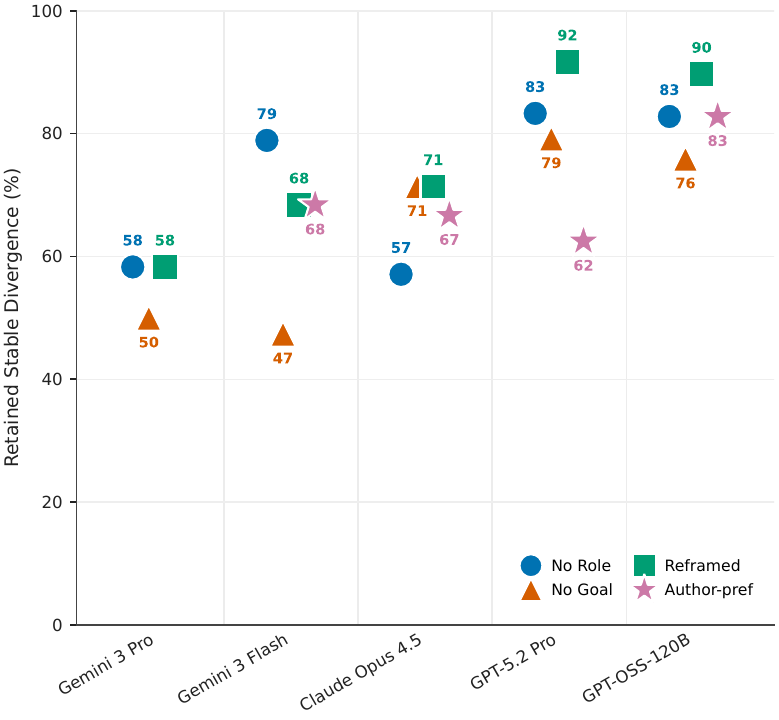}
    \caption{Retained-divergence rates for selected models across prompt and author-preference ablations.}
    \label{fig:retention_ablation_profiles}
\end{subfigure}
\caption{Robustness and agreement structure of model future-work choices: inter-model agreement (left) and retained stable-divergence rates under prompt and author-preference variations (right).}
\label{fig:robustness_and_agreement}
\end{figure}

Across all four prompt variations, models retained the majority of their stable divergences with weighted mean retained divergences across all models ranging from 66.9\% to 80.5\%. Figure~\ref{fig:retention_ablation_profiles} shows retained-divergence rates for selected models across all four conditions. When divergences did not hold, they predominantly became unstable choices across orderings rather than correcting toward preference for author-recommended directions, suggesting that prompt constructions contributed to choice consistency more than to the direction of the preference itself. Appendix Tables~\ref{tab:prompt_variance} and~\ref{tab:authorprefpred_stable_divergence} include the unstable choice and reversal to author alignment rates for each model. Appendix Table~\ref{tab:model_retention_ci} reports model-specific Wilson confidence intervals for retained stable divergence rates.

Per-model behavior varied substantially across the four prompt variations. Gemini 3 Pro and Gemini 3 Flash were especially sensitive to prompt changes, specifically the no-goal variation. Divergence rates dropped to 50\% and 47\%, respectively, and unstable choices across the two orderings increased to 42\% and 37\%, respectively, suggesting that their canonical divergences are partly anchored by research goal interpretation. On the author preference prediction condition, reversals are somewhat concentrated in GPT-5.2 Pro and Claude Opus 4.5, where this framing produces higher reversal rates than any of the ordinary prompt ablations for the same models. Figure~\ref{fig:reasoning_reversal} illustrates one such reversal.

\subsection{How do humans do on this task?}

We also report a pilot human baseline study on the same task. We collected 108 responses from 10 participants on an anonymized, filtered subset of 23 of the 80 unique query pairs. Unlike the models, each human participant saw only one randomized ordering of each option pair, could complete any number of queries, and could abstain if they thought the query was underdetermined due to insufficient context, unclear options, or a lack of immediate domain expertise. Response coverage across the 23 query pairs ranges from 1 to 8 substantive judgments per query, with only three queries receiving fewer than three substantive responses. Per-query response counts, abstention reasons, and participant qualifications are reported in Appendix tables~\ref{tab:human_per_pair} and~\ref{tab:human_participants}.

Of the 108 responses, 13 (12.0\%) were abstentions. Among the remaining 95 fixed-choice responses, 58 (61.1\%) selected the author-recommended direction and 37 (38.9\%) selected the sampled distractor. Because participants completed different numbers of queries, we also computed per-query macro-averaged choice rates by first computing the choice rate for each query, excluding abstentions, and then averaging across all 23 queries. These were similar to the pooled result, with 61.6\% author-aligned choices and 38.4\% alternative choices. Majority-vote analyses across query responses also show the same qualitative pattern. Full results are shared in Appendix table~\ref{tab:human_aggregate} including subset analyses restricted to pairs with at least three fixed-choice responses, and pairs with no query-quality abstentions.

Overall, the pilot suggests that participants can engage with the task, as reflected in the relatively low abstention rate, while also showing non-trivial variation across responses with a modest overall preference for author-recommended directions over plausible alternatives. Because human participants saw only one ordering of each pair, we do not directly compare these responses to the model author-alignment, stable-divergence, and unstable-choice metrics. 

\subsection{What taste patterns emerge from stable divergences?}
\label{qual_analysis}

To further analyze the structure of these 169 elicited stable divergence events across 45 unique queries, we labeled both the author-recommended future direction and the extracted distractor according to the type of extension they proposed relative to the project's research goal. We used a five-category taxonomy: Methodological/Architectural, Infrastructure/Data, Scaling/Generalization, Application/Downstream, and Mechanistic/Interpretability. The taxonomy and category definitions are provided in Appendix table~\ref{tab:taxonomy}.

We next analyzed the author-to-distractor category transitions represented across the 169 stable-divergence events. The complete transition table is reported in Appendix Table~\ref{tab:transitions}, which shows both each transition’s share of all stable-divergence events and, within each author-recommended category, the distribution of selected distractor categories. We also include a smaller selected view of the most frequent transitions in the 144-event subset from pairs where at least three models diverged in Appendix table~\ref{tab:transitions_selected}. These patterns remain broadly similar to the transition distribution in the full 169-event set. 

Of the 169 events, 128 (75.7\%) crossed category boundaries, while 41 (24.3\%) remained within the same broad category. These within-category events underscore the coarseness of our early five-category taxonomy. Two meaningfully different research directions may still propose the same high-level type of extension to the research project. We include examples of both within-category and cross-category transition labels in the Appendix as well. 

Across the 128 cross-category events, the three most frequent transition families were Infrastructure-to-Methodological substitutions (55 events), Methodological-to-Scaling transitions (23 events), and Application-to-Scaling transitions (12 events). Together, these families account for 90 of the 128 cross-category events (70.3\%), or 53.3\% of all stable-divergence events. 

The Infrastructure-to-Methodological pattern is bidirectional and reflects a recurring trade-off between these types of future research extensions, whereas the latter two patterns show a clearer directional movement toward scaling and generalization. Because this analysis includes only stable divergence events and does not account for how frequently each transition was available, we report these counts descriptively rather than as normative judgments of model-level preferences. To illustrate the qualitative character of these divergences, we present representative reasoning traces across selected transition categories in Appendix figures~\ref{fig:reasoning_reversal},~\ref{fig:reasoning_scaling} and~\ref{fig:reasoning_infra_method}.

\section{Related Works}
\label{related_works}

\textbf{LLM-Driven Idea Generation and Scientific Capability Benchmarks:} 
Recent survey work maps the growing literature on hypothesis generation using large language models \citep{alkan2025surveyhypothesisgenerationscientific, kulkarni2025scientifichypothesisgenerationvalidation}. This literature includes data-driven hypothesis generation, where models propose and verify hypotheses within a dataset \citep{liu2025hypobenchsystematicprincipledbenchmarking}, and literature-driven hypothesis generation, where models combine ideas from existing papers. More broadly, scientific capability benchmarks evaluate tasks such as hypothesis generation, insight synthesis, and scientific reasoning. Our task differs in focus. We do not primarily ask whether a model can generate a correct hypothesis or even rank candidates against a single best answer as in \citet{liu2025researchbenchbenchmarkingllmsscientific}. Instead, we study preference under plausible alternatives. Each query compares an author-recommended future direction against a distinct but plausible distractor sampled from a related paper. 

\textbf{Falsification \& Execution-Grounded Research Automation:}
A complementary line of work studies AI systems for scientific discovery when hypotheses or research ideas can be tested against executable or falsifiable objectives. \citet{huang2025automatedhypothesisvalidationagentic} validate free-form hypotheses by having LLM agents design falsification experiments under strict Type-I error control, while \citet{si2026executiongroundedautomatedairesearch} use execution feedback from GPU experiments to improve LLM-generated research ideas. Related execution-grounded evaluators inspect code and data artifacts to assess reproducibility and generalizability of scientific claims \citep{bai2026storynotscience}. These systems highlight the power of verifiability and execution feedback, but they are most directly applicable once a research problem has a checkable objective and feasible execution path. Our future-work choice task targets open-ended scientific problems where execution of multiple hypotheses might be constrained, or no single, clear measure for progress can be determined.

\textbf{Scientific Impact Forecasting:}
A parallel line of work evaluates whether language models can forecast scientific outcomes, including empirical results \citep{wen2025predictingempiricalairesearch}, the relative promise of competing ideas \citep{ye2026prooftimebenchmarkevaluating}, and which scientific contributions will ultimately materialize \citep{ajith2026presciencebenchmarkforecastingscientific, wu2026forecastingscientificprogressartificial}. We view these tasks as adjacent because forecasting scientific impact can elicit signals related to judgment, foresight, and research preference. However, forecasting asks models to predict downstream outcomes, while our task asks models to choose a next research direction in a situated project context. Our author-preference framing check also suggests that next-step choice and explicit author-preference prediction need not elicit identical behavior, motivating future comparisons between impact forecasting capabilities and agentic next-step selection.

\textbf{Existing Research Taste Literature:}
Most directly related is a nascent line of work that attempts to measure or instill research taste in language models. \citet{mahajan2025tastybench} measure research taste in judge models using citation velocity as a proxy for high taste. \citet{tong2026ailearnscientifictaste} extend this to include other large-scale community signals to train a judge model and use it for training idea-generator LMs. \citet{gong2026llmslearnscientifictaste} additionally use institutional traces to finetune models toward high-taste research judgments. We place our work in the same problem space, the lack of clear learning and evaluation signals for what these papers describe as ``strategic taste,'' ``strong judgment and foresight for impact,'' and ``evaluative judgment in low-verifiability domains.'' However, our goal is not to score models as having high or low taste, or to train a judge model. Instead, we begin to characterize descriptive taste profiles through one situated behavioral signal of choosing a research direction from between two plausible alternatives given the same research context.

\section{Limitations \& Future Work}
\label{sec:limitations_and_future_work}

\textbf{Construct Validity \& Interpretation}. Ground truth reflects the author's preference, not an objectively optimal research direction. Distractors are real future works from related papers and could genuinely be better directions. We therefore frame results as author alignment rather than accuracy, and focus on characterizing taste profiles rather than evaluating correctness or recurring failure modes. Post-cutoff contamination also cannot be ruled out entirely, especially for API-accessed models whose training data or retrieval access may extend beyond stated cutoff dates. Future versions should ground human research-direction preferences in richer traces beyond future-work statements, such as author-specific citation trees and expert-sourced steering trajectories over research directions.

\textbf{Dataset Construction Rigor}. All extractions are performed using a single model (Claude Sonnet 4.5), which is also used as the LLM-as-judge for distractor filtering. Using a single model across extraction, rephrasing, and judging can amplify any biases it carries into the final query set. Although we conduct manual author review of query quality, this audit is limited. Future versions should include more rigorous query-level quality control, including multi-model LLM-as-judge scoring and expert human ratings of whether each prompt with its context and options forms a sufficient, fair and meaningful research-direction choice.

\textbf{Dataset Coverage}. The dataset comprises 80 pairwise comparisons from 19 papers across ML subdomains, so we cannot claim generalization to other scientific fields or broader coverage within ML. Future work should scale up the dataset within ML and extend the task to scientific domains beyond ML, such as biology, physics, and the social sciences, to test domain-specific preferences and emergence of cross-domain model preferences and/or metascientific reasoning.

\textbf{Analysis Scope \& Extensions}. The transition analysis is conducted only on the stable-divergence subset, so it characterizes the structure of observed disagreement rather than the relative likelihood of divergence versus author alignment for each transition type. Future work should compare this selection-based probe with free-form generation of future research directions, and test how inference settings such as temperature, the form of context provided, and different degrees of agency affect taste consistency. To better situate these behaviors relative to the human scientist community, future work could also scale the pilot human baseline to larger expert and non-expert cohorts for a fuller comparison of human and LLM research-direction preferences.

\clearpage
\bibliographystyle{colm2026_conference}
\bibliography{references}

@misc{ajith2026presciencebenchmarkforecastingscientific,
      title={{PreScience}: A Benchmark for Forecasting Scientific Contributions}, 
      author={Anirudh Ajith and Amanpreet Singh and Jay DeYoung and Nadav Kunievsky and Austin C. Kozlowski and Oyvind Tafjord and James Evans and Daniel S. Weld and Tom Hope and Doug Downey},
      year={2026},
      eprint={2602.20459},
      archivePrefix={arXiv},
      primaryClass={cs.AI},
      url={https://arxiv.org/abs/2602.20459}, 
}

@misc{alkan2025surveyhypothesisgenerationscientific,
      title={A Survey on Hypothesis Generation for Scientific Discovery in the Era of Large Language Models}, 
      author={Atilla Kaan Alkan and Shashwat Sourav and Maja Jablonska and Simone Astarita and Rishabh Chakrabarty and Nikhil Garuda and Pranav Khetarpal and Maciej Pióro and Dimitrios Tanoglidis and Kartheik G. Iyer and Mugdha S. Polimera and Michael J. Smith and Tirthankar Ghosal and Marc Huertas-Company and Sandor Kruk and Kevin Schawinski and Ioana Ciucă},
      year={2025},
      eprint={2504.05496},
      archivePrefix={arXiv},
      primaryClass={cs.CL},
      url={https://arxiv.org/abs/2504.05496}, 
}

@misc{bai2026storynotscience,
      title={The Story is Not the Science: Execution-Grounded Evaluation of Mechanistic Interpretability Research},
      author={Xiaoyan Bai and Alexander Baumgartner and Haojia Sun and Ari Holtzman and Chenhao Tan},
      year={2026},
      eprint={2602.18458},
      archivePrefix={arXiv},
      primaryClass={cs.CY},
      url={https://arxiv.org/abs/2602.18458},
}

@article{
doi:10.1126/science.aao0185,
author = {Santo Fortunato  and Carl T. Bergstrom  and Katy Börner  and James A. Evans  and Dirk Helbing  and Staša Milojević  and Alexander M. Petersen  and Filippo Radicchi  and Roberta Sinatra  and Brian Uzzi  and Alessandro Vespignani  and Ludo Waltman  and Dashun Wang  and Albert-László Barabási },
title = {Science of science},
journal = {Science},
volume = {359},
number = {6379},
pages = {eaao0185},
year = {2018},
doi = {10.1126/science.aao0185},
URL = {https://www.science.org/doi/abs/10.1126/science.aao0185},
eprint = {https://www.science.org/doi/pdf/10.1126/science.aao0185}}

@misc{gong2026llmslearnscientifictaste,
      title={{LLMs} learn scientific taste from institutional traces across the social sciences},
      author={Ziqin Gong and Ning Li and Huaikang Zhou},
      year={2026},
      eprint={2603.16659},
      archivePrefix={arXiv},
      primaryClass={cs.AI},
      url={https://arxiv.org/abs/2603.16659},
}

@misc{huang2025automatedhypothesisvalidationagentic,
      title={Automated Hypothesis Validation with Agentic Sequential Falsifications}, 
      author={Kexin Huang and Ying Jin and Ryan Li and Michael Y. Li and Emmanuel Candès and Jure Leskovec},
      year={2025},
      eprint={2502.09858},
      archivePrefix={arXiv},
      primaryClass={cs.LG},
      url={https://arxiv.org/abs/2502.09858}, 
}

@misc{huggingface2026mlintern,
      title={ml-intern: an open-source {ML} engineer that reads papers, trains models, and ships {ML} models},
      author={{Hugging Face}},
      year={2026},
      howpublished={\url{https://github.com/huggingface/ml-intern}},
      note={GitHub repository},
}

@misc{karpathy2026autoresearch,
      title={autoresearch: {AI} agents running research on single-{GPU} nanochat training automatically},
      author={Andrej Karpathy},
      year={2026},
      howpublished={\url{https://github.com/karpathy/autoresearch}},
      note={GitHub repository},
}

@misc{kulikov2026autodata,
      title={Autodata: An agentic data scientist to create high quality synthetic data},
      author={Ilia Kulikov and Chenxi Whitehouse and Tianhao Wu and Yixin Nie and Swarnadeep Saha and Eryk Helenowski and Weizhe Yuan and Olga Golovneva and Jack Lanchantin and Yoram Bachrach and Jakob Foerster and Xian Li and Han Fang and Sainbayar Sukhbaatar and Jason Weston},
      year={2026},
      eprint={2606.25996},
      archivePrefix={arXiv},
      primaryClass={cs.AI},
      url={https://arxiv.org/abs/2606.25996},
}

@misc{kulkarni2025scientifichypothesisgenerationvalidation,
      title={Scientific Hypothesis Generation and Validation: Methods, Datasets, and Future Directions}, 
      author={Adithya Kulkarni and Fatimah Alotaibi and Xinyue Zeng and Longfeng Wu and Tong Zeng and Barry Menglong Yao and Minqian Liu and Shuaicheng Zhang and Lifu Huang and Dawei Zhou},
      year={2025},
      eprint={2505.04651},
      archivePrefix={arXiv},
      primaryClass={cs.CL},
      url={https://arxiv.org/abs/2505.04651}, 
}

@misc{liu2025hypobenchsystematicprincipledbenchmarking,
      title={{HypoBench}: Towards Systematic and Principled Benchmarking for Hypothesis Generation}, 
      author={Haokun Liu and Sicong Huang and Jingyu Hu and Yangqiaoyu Zhou and Chenhao Tan},
      year={2025},
      eprint={2504.11524},
      archivePrefix={arXiv},
      primaryClass={cs.AI},
      url={https://arxiv.org/abs/2504.11524}, 
}

@misc{liu2025researchbenchbenchmarkingllmsscientific,
      title={{ResearchBench}: Benchmarking {LLMs} in Scientific Discovery via Inspiration-Based Task Decomposition}, 
      author={Yujie Liu and Zonglin Yang and Tong Xie and Jinjie Ni and Ben Gao and Yuqiang Li and Shixiang Tang and Wanli Ouyang and Erik Cambria and Dongzhan Zhou},
      year={2025},
      eprint={2503.21248},
      archivePrefix={arXiv},
      primaryClass={cs.CL},
      url={https://arxiv.org/abs/2503.21248}, 
}

@misc{mahajan2025tastybench,
  title = {{TastyBench}: Toward Measuring Research Taste in {LLM}},
  author = {Mahajan, Parv and Yilin and yix},
  year = {2025},
  month = dec,
  url = {https://www.lesswrong.com/posts/Mxsy7wYvsCRv5dGrw/tastybench-toward-measuring-research-taste-in-llm},
  note = {LessWrong post}
}

@misc{novikov2025alphaevolvecodingagentscientific,
      title={{AlphaEvolve}: A coding agent for scientific and algorithmic discovery},
      author={Alexander Novikov and Ngân Vũ and Marvin Eisenberger and Emilien Dupont and Po-Sen Huang and Adam Zsolt Wagner and Sergey Shirobokov and Borislav Kozlovskii and Francisco J. R. Ruiz and Abbas Mehrabian and M. Pawan Kumar and Abigail See and Swarat Chaudhuri and George Holland and Alex Davies and Sebastian Nowozin and Pushmeet Kohli and Matej Balog},
      year={2025},
      eprint={2506.13131},
      archivePrefix={arXiv},
      primaryClass={cs.AI},
      url={https://arxiv.org/abs/2506.13131},
}

@misc{sharma2025openevolve,
      title={{OpenEvolve}: an open-source evolutionary coding agent},
      author={Asankhaya Sharma},
      year={2025},
      howpublished={\url{https://github.com/algorithmicsuperintelligence/openevolve}},
      note={GitHub repository},
}

@misc{si2026executiongroundedautomatedairesearch,
      title={Towards Execution-Grounded Automated {AI} Research}, 
      author={Chenglei Si and Zitong Yang and Yejin Choi and Emmanuel Candès and Diyi Yang and Tatsunori Hashimoto},
      year={2026},
      eprint={2601.14525},
      archivePrefix={arXiv},
      primaryClass={cs.CL},
      url={https://arxiv.org/abs/2601.14525}, 
}

@misc{tong2026ailearnscientifictaste,
      title={{AI} Can Learn Scientific Taste},
      author={Jingqi Tong and Mingzhe Li and Hangcheng Li and Yongzhuo Yang and Yurong Mou and Weijie Ma and Zhiheng Xi and Hongji Chen and Xiaoran Liu and Qinyuan Cheng and Ming Zhang and Qiguang Chen and Weifeng Ge and Qipeng Guo and Tianlei Ying and Tianxiang Sun and Yining Zheng and Xinchi Chen and Jun Zhao and Ning Ding and Xuanjing Huang and Yugang Jiang and Xipeng Qiu},
      year={2026},
      eprint={2603.14473},
      archivePrefix={arXiv},
      primaryClass={cs.CL},
      url={https://arxiv.org/abs/2603.14473},
}

@misc{wen2025predictingempiricalairesearch,
      title={Predicting Empirical {AI} Research Outcomes with Language Models}, 
      author={Jiaxin Wen and Chenglei Si and {Yueh-han} Chen and He He and Shi Feng},
      year={2025},
      eprint={2506.00794},
      archivePrefix={arXiv},
      primaryClass={cs.AI},
      url={https://arxiv.org/abs/2506.00794}, 
}

@misc{wu2026forecastingscientificprogressartificial,
      title={Forecasting Scientific Progress with Artificial Intelligence}, 
      author={Sean Wu and Pan Lu and Yupeng Chen and Jonathan Bragg and Yutaro Yamada and Peter Clark and David Clifton and Philip Torr and James Zou and Junchi Yu},
      year={2026},
      eprint={2605.22681},
      archivePrefix={arXiv},
      primaryClass={cs.AI},
      url={https://arxiv.org/abs/2605.22681}, 
}

@misc{ye2026prooftimebenchmarkevaluating,
      title={Proof of Time: A Benchmark for Evaluating Scientific Idea Judgments}, 
      author={Bingyang Ye and Shan Chen and Jingxuan Tu and Chen Liu and Zidi Xiong and Samuel Schmidgall and Danielle S. Bitterman},
      year={2026},
      eprint={2601.07606},
      archivePrefix={arXiv},
      primaryClass={cs.CL},
      url={https://arxiv.org/abs/2601.07606}, 
}

@misc{zhang2025darwingodelmachine,
      title={{Darwin G{\"o}del Machine}: Open-Ended Evolution of Self-Improving Agents},
      author={Jenny Zhang and Shengran Hu and Cong Lu and Robert Lange and Jeff Clune},
      year={2025},
      eprint={2505.22954},
      archivePrefix={arXiv},
      primaryClass={cs.AI},
      url={https://arxiv.org/abs/2505.22954},
}

@misc{zheng2024largelanguagemodelsrobust,
      title={Large Language Models Are Not Robust Multiple Choice Selectors}, 
      author={Chujie Zheng and Hao Zhou and Fandong Meng and Jie Zhou and Minlie Huang},
      year={2024},
      eprint={2309.03882},
      archivePrefix={arXiv},
      primaryClass={cs.CL},
      url={https://arxiv.org/abs/2309.03882}, 
}
\clearpage
\appendix

\section{Qualitative Reasoning Examples}
\label{appendix:examples}

\begin{figure}[h]
\begin{tcolorbox}[exhibitbase, breakable=false, fontupper=\footnotesize, colback=panelfill, colframe=gray!55]
\begin{tcolorbox}[respinner, title={The choice}]
A multi-agent ``scientific assistant'' framework reports strong single-task results (90\% agent-routing and 97.5\% task-completion on synthetic benchmarks, comparable to frontier models on a chemistry benchmark), with its weakest component a simulation agent at 75\% routing. 
\end{tcolorbox}
\begin{tcbraster}[raster columns=2, raster equal height, raster column skip=8pt, raster left skip=0pt, raster right skip=0pt]
\begin{tcolorbox}[respinner, title={Canonical Prompt}]
\textbf{B (distractor):} ``Investigate more robust constraint mechanisms to effectively guide agents through long-horizon, repetitive tasks.''\par\smallskip
\itshape ``Current 90\% routing accuracy \textbf{\upshape compounds poorly over multiple steps} (0.9\textasciicircum10 $\approx$ 35\% success for 10-step workflows) \ldots{} Before expanding infrastructure integration (Option A), \textbf{\upshape the framework should first demonstrate it can reliably maintain coherence} and complete complex, multi-step scientific workflows.''
\end{tcolorbox}
\begin{tcolorbox}[respauthor, title={Predicting Author Preference}]
\textbf{A (author):} ``Explore customized orchestration backends that support richer control flow and tighter integration with scientific compute environments.''\par\smallskip
\itshape ``The most notable weakness explicitly mentioned is the \textbf{\upshape simulation agent's 75\% routing accuracy} \ldots{} improving orchestration backends \ldots{} would directly address their documented limitation \ldots{} \textbf{\upshape Long-horizon task limitations were not explicitly identified in the results}.''
\end{tcolorbox}
\end{tcbraster}
\end{tcolorbox}
\caption{An author-preference reversal excerpts from Opus~4.5. With the canonical prompt, model focuses on future capability and picks the distractor. Asked to predict the \emph{authors'} likely choice, it anchors on the one limited result number in the context and selects the author recommended direction to improve upon it.}
\label{fig:reasoning_reversal}
\end{figure}

\begin{figure}[h]
\begin{tcolorbox}[exhibitbase, breakable=false, fontupper=\footnotesize, colback=panelfill, colframe=gray!55]
\begin{tcolorbox}[respinner, title={The choice}]
A prefix-tuning method reports strong results on math and scientific reasoning benchmarks (matching or surpassing weight-based fine-tuning while using 59\% less data and 46\% fewer parameters), with the pretrained model weights kept frozen.
\end{tcolorbox}
\begin{tcbraster}[raster columns=2, raster equal height, raster column skip=8pt, raster left skip=0pt, raster right skip=0pt]
\begin{tcolorbox}[respinner, title={GPT-5.2 Pro\hfill\textnormal{Scaling}}]
\textbf{B (distractor):} ``Explore methods that explicitly test generalizability and scalability, and incorporate them as optimization objectives.''\par\smallskip
\itshape ``Direction A \ldots{}primarily converts the already-demonstrated method into \textbf{\upshape engineering leverage; it doesn't most directly strengthen the central research claim} \ldots{} Direction B directly addresses the next most important research question: \textbf{\upshape whether the learned representations produce reliable, scalable generalization} \ldots{}''
\end{tcolorbox}
\begin{tcolorbox}[respauthor, title={Gemini 3 Pro\hfill\textnormal{Application}}]
\textbf{A (author):} ``Integrate trainable cache representations into existing inference systems and post-training frameworks \ldots{}''\par\smallskip
\itshape ``The most practical and impactful next step is to \textbf{\upshape deploy this mechanism} \ldots{} Integrating these trainable representations into existing inference engines and post-training pipelines will allow the research to \textbf{\upshape demonstrate real-world utility} \ldots{}''
\end{tcolorbox}
\end{tcbraster}
\end{tcolorbox}
\caption{An Application to Scaling divergence between GPT-5.2 Pro and Gemini 3 Pro. GPT-5.2 Pro treats the author's deployment direction as engineering and picks the distractor to test the method's generalization. Gemini 3 Pro selects the author's direction, prioritizing real-world deployment of the proven efficiency gains. GPT-OSS 120B agrees with GPT-5.2 Pro, calling deployment an engineering effort rather than a test of generalization.}
\label{fig:reasoning_scaling}
\end{figure}

\begin{figure}[h]
\begin{tcolorbox}[exhibitbase, breakable=false, fontupper=\footnotesize, colback=panelfill, colframe=gray!55]
\begin{tcolorbox}[respinner, title={The choice}]
A system for generating and executing research ideas, trained with RL from execution reward, reports a high aggregate execution rate (over 90\%) and gains in average reward, with RL failing to raise the \emph{maximum} reward (H6, invalidated) and collapsing idea diversity toward simple ideas, where ideas with longer reasoning show consistently lower execution rates (H7).
\end{tcolorbox}
\begin{tcbraster}[raster columns=2, raster equal height, raster column skip=8pt, raster left skip=0pt, raster right skip=0pt]
\begin{tcolorbox}[respinner, title={Opus 4.5\hfill\textnormal{Methodological}}]
\textbf{B (distractor):} ``Investigate designed verification tasks, more principled integration of verification and generation objectives \ldots{}''\par\smallskip
\itshape ``The critical failure is that RL doesn't improve upper-bound performance and causes diversity collapse. Option B directly targets this core problem, while \textbf{\upshape Option A addresses a secondary concern (execution capability) that already shows strong performance (90\%+ rates)} \ldots{} how can we make RL actually improve the \textbf{\upshape best} ideas generated, not just the average?''
\end{tcolorbox}
\begin{tcolorbox}[respauthor, title={GPT-OSS 120B\hfill\textnormal{Infrastructure}}]
\textbf{A (author):} ``Improve the execution agent to enable correct implementation of more complex idea types \ldots{}''\par\smallskip
\itshape ``The execution agent already achieves very high success rates for many ideas, but RL collapses to simple ideas \textbf{\upshape because more complex, higher-impact ideas cannot be reliably executed} \ldots{} enhancing the execution agent so it can implement more sophisticated ideas would unlock a larger space of high-potential ideas.''
\end{tcolorbox}
\end{tcbraster}
\end{tcolorbox}
\caption{An Infrastructure to Methodological divergence between Opus 4.5 and GPT-OSS 120B reading the same execution-rate evidence. Opus 4.5 reads the high \emph{aggregate} execution rate as evidence that execution is solved and picks the distractor to fix the learning objective. GPT-OSS 120B reads the \emph{lower} rate on complex ideas as the binding constraint and selects the author's executor-focused direction.}
\label{fig:reasoning_infra_method}
\end{figure}

\clearpage
\section{Prompts}
\label{appendix:prompts}

\subsubsection{Paper Triage Prompt}
\label{prompt:triage}

\begin{promptbox}
Does this paper have explicitly stated recommendations for future work and next steps?
Here is the LaTeX source of the paper:
{tex_content}
\end{promptbox}

\subsubsection{Research Goal Extraction Prompt}
\label{prompt:research_goal}

\begin{promptbox}
Extract one high-level research goal from this paper that covers all subgoals.

Provide:
1. The research goal in your own words, synthesized from the paper.
2. Key supporting statements from the paper with their sections.

Output in the following format:
<research_goal>
<text>...</text>
<citations>...</citations>
</research_goal>

Output plain text only - convert any LaTeX commands or macros (e.g., \emph{}, \textbf{}, \ours{}) to their intended meaning.

Here is the LaTeX source of the paper:
{tex_content}
\end{promptbox}

\subsubsection{Hypotheses Tested \& Key Results Extraction Prompt}
\label{prompt:hypotheses}

\begin{promptbox}
Extract the main hypotheses tested and their key results from this paper.

For each hypothesis, provide:
1. The hypothesis or main claim (exact text, preserving original wording)
2. The key result or finding that validates/invalidates it
3. The section it came from

Make sure each hypothesis is substantial, not a singular sub-claim or ablation.

If no explicit hypothesis is stated, extract the main claims validated through experiments.

Output in the following format:
<hypotheses_tested>
<item num="1">
<text>...</text>
<key_result>...</key_result>
<citation>...</citation>
</item>
</hypotheses_tested>

Output plain text only - convert any LaTeX commands or macros (e.g., \emph{}, \textbf{}, \ours{}) to their intended meaning.

Here is the LaTeX source of the paper:
{tex_content}
\end{promptbox}

\subsubsection{Future Works Extraction Prompt}
\label{prompt:future_works}

\begin{promptbox}
Extract all future work recommendations and next steps from this paper.

For each recommendation, provide:
1. The exact text from the paper preserving original wording.
2. The section it came from with relevant position pointers.

Only include explicitly stated future work, not implied directions.

Output the future works as a number list in the following format:
<future_works>
<item num="1">
<extracted_text>...</extracted_text>
<citation>...</citation>
</item>
</future_works>

Here is the LaTeX source of the paper:
{tex_content}
\end{promptbox}

\subsubsection{Future Works Rephrasing Prompt}
\label{prompt:rephrasing}

\begin{promptbox}
Rephrase the following future work recommendation from a paper into a clear, actionable research direction statement.

The output should be a single sentence describing what to do or investigate. Focus on the proposed action(s).

Keep the core idea intact. Do not add information not present in the original text shared with you. Preserve the original phrasing and terminology where possible.

In case the future work text includes specific names of methods, datasets, architectures, or benchmarks, avoid those and keep the statement general and focused on the research direction. Describe the specifics by their type or purpose.

Output in the following format:
<research_direction_statement>
...
</research_direction_statement>

Here is the high level research goal of the paper to provide context:
{research_goal}

Here is the future work text to rephrase:
{fw_text}
\end{promptbox}

\subsubsection{Distractor Evaluation Prompt (LLM-as-Judge)}
\label{prompt:distractor_eval}

\begin{promptbox}
Evaluate whether a candidate research direction can serve as a good distractor in a multiple-choice question about selecting the next research hypothesis.

A good distractor must satisfy two criteria:
1. DISTINCT: The candidate must be meaningfully different from all ground truth future works. It should not overlap semantically with any of them, even if worded differently.
2. PLAUSIBLE: The candidate must be a reasonable next step given the research goal. It should be topically relevant and methodologically compatible with the paper's direction. It should not be an obviously inferior or regressive step compared to the paper's current state.

First, reason through each criterion step by step. Then provide binary judgments.

Output in the following format:
<distractor_evaluation>
<distinctness_reasoning>...</distinctness_reasoning>
<plausibility_reasoning>...</plausibility_reasoning>
<is_distinct>yes/no</is_distinct>
<is_plausible>yes/no</is_plausible>
</distractor_evaluation>

Here is the target paper's research goal:
{research_goal}

Here are the target paper's ground truth future works (rephrased as research direction statements):
{ground_truth_future_works}

Here is the candidate distractor to evaluate:
{distractor_candidate}
\end{promptbox}

\subsubsection{Future-Work Selection Prompt (Evaluation)}
\label{prompt:evaluation}

\begin{promptbox}
You are an AI research assistant, currently helping decide the next research direction for a running research project. Below is all the context you need. Think step by step through the information provided before answering.

## Research Goal
{research_goal}

## Hypotheses Already Tested & Key Results
1. **Hypothesis**: {hypothesis}
   **Key Result**: {key_result}
...

## Which research direction should this project pursue next?

## Your response must follow this format:

Thought Process: <your step by step reasoning here>
Final Answer: <one character: A or B>

A. {option A}

B. {option B}
\end{promptbox}

\subsubsection{Prompt Variation Conditions}
\label{prompt:variations}
Each condition changes a single element of the evaluation prompt (Appendix~\ref{prompt:evaluation}); all other elements are held identical. \textbf{No Research Goal} removes the \texttt{\#\# Research Goal} section. The other three replace one line, as shown below.

\noindent\textbf{No Role}
\begin{promptbox}
Below is all the context you need. Think step by step through the information provided before answering.
\end{promptbox}

\noindent\textbf{Reframed Task}
\begin{promptbox}
Which option is the better next move for this project?
\end{promptbox}

\noindent\textbf{Author Preference Prediction}
\begin{promptbox}
You are an AI research assistant, currently helping predict which future research direction the researchers behind this work would choose next for their own project. Below is all the context you need. Think step by step through the information provided before answering.

Which research direction would the researchers behind this work be more likely to choose for this project?
\end{promptbox}

\clearpage
\section{Additional Results and Tables}
\label{appendix:additional_results}

\begin{table}[ht]
    \centering
    \caption{Average semantic (embedding) and lexical (TF-IDF) cosine similarity across 80 unique query pairs.}
    \label{tab:similarity_controls}
    \begin{tabular}{lcc}
    \toprule
    \textbf{Pair} & \textbf{Embedding} & \textbf{TF-IDF} \\
    \midrule
    GT, Distractor & 0.41 & 0.05 \\
    GT, Context & 0.48 & 0.14 \\
    Distractor, Context & 0.40 & 0.08 \\
    $\Delta$(GT, Context) $-$ (Dist, Context) & 0.08 $\pm$ 0.09 & 0.05 $\pm$ 0.07 \\
    \bottomrule
    \end{tabular}
\end{table}

\begin{table}[htbp]
    \centering
    \caption{Pairwise future-work selection results (n=80 pairs, 160 ordered queries). Cells report percentages with Wilson 95\% confidence intervals in brackets; each outcome uses 80 pairwise comparisons as the denominator.}
    \label{tab:pairwise_results}
    \begin{tabular}{llccc}
    \toprule
    \textbf{Family} & \textbf{Model} & \textbf{Author Alignment} & \textbf{Stable Divergence} & \textbf{Unstable Choice} \\
    \midrule
    \multirow{2}{*}{Google}
     & Gemini 3 Pro & \rateci{67.5}{56.6}{76.8} & \rateci{15.0}{8.8}{24.4} & \rateci{17.5}{10.7}{27.3} \\
     & Gemini 3 Flash & \rateci{62.5}{51.5}{72.3} & \rateci{23.8}{15.8}{34.1} & \rateci{13.8}{7.9}{23.0} \\
    \midrule
    \multirow{2}{*}{Anthropic}
     & Claude Opus 4.5 & \rateci{55.0}{44.1}{65.4} & \rateci{26.2}{17.9}{36.8} & \rateci{18.8}{11.7}{28.7} \\
     & Claude Haiku 4.5 & \rateci{50.0}{39.3}{60.7} & \rateci{21.2}{13.7}{31.4} & \rateci{28.7}{20.0}{39.5} \\
    \midrule
    \multirow{2}{*}{OpenAI}
     & GPT-5.2 Pro & \rateci{57.5}{46.6}{67.7} & \rateci{30.0}{21.1}{40.8} & \rateci{12.5}{6.9}{21.5} \\
     & GPT-OSS-120B & \rateci{50.0}{39.3}{60.7} & \rateci{36.2}{26.6}{47.2} & \rateci{13.8}{7.9}{23.0} \\
    \midrule
    Qwen & Qwen3 Max Think. & \rateci{50.0}{39.3}{60.7} & \rateci{23.8}{15.8}{34.1} & \rateci{26.2}{17.9}{36.8} \\
    \midrule
    Meta & Llama 4 Maverick & \rateci{45.0}{34.6}{55.9} & \rateci{35.0}{25.5}{45.9} & \rateci{20.0}{12.7}{30.0} \\
    \midrule
    \multicolumn{2}{l}{\textit{Random Baseline}} & 25.0\% & 25.0\% & 50.0\% \\
    \bottomrule
    \end{tabular}
    \end{table}

\begin{table}[htbp]
\centering
\caption{Cumulative stable divergence prevalence across 80 pairwise comparisons at each model-count threshold.}
\label{tab:divergence_prevalence}
\begin{tabular}{lc}
\toprule
\textbf{Threshold} & \textbf{Pairs} \\
\midrule
$\geq$ 1 model & 45/80 \\
$\geq$ 2 models & 34/80 \\
$\geq$ 3 models & 27/80 \\
$\geq$ 4 models & 24/80 \\
$\geq$ 5 models & 18/80 \\
$\geq$ 6 models & 11/80 \\
$\geq$ 7 models & 7/80 \\
$\geq$ 8 models & 3/80 \\
\bottomrule
\end{tabular}
\end{table}

\begin{table}[htbp]
\centering
\caption{Stable divergence pairs by similarity bucket (embedding). GT More Similar denotes pairs where models preferred the distractor despite GT being more similar to context.}
\label{tab:divergence_overlap}
\begin{tabular}{lccc}
\toprule
\textbf{Model} & \textbf{Stable Div.} & \textbf{GT More Similar} & \textbf{Dist More Similar} \\
\midrule
Google Gemini 3 Pro & 12 & 8 (66.7\%) & 4 (33.3\%) \\
Google Gemini 3 Flash & 19 & 14 (73.7\%) & 5 (26.3\%) \\
Claude Opus 4.5 & 21 & 16 (76.2\%) & 5 (23.8\%) \\
Claude Haiku 4.5 & 17 & 14 (82.4\%) & 3 (17.6\%) \\
GPT-5.2 Pro & 24 & 18 (75.0\%) & 6 (25.0\%) \\
GPT-OSS-120B & 29 & 23 (79.3\%) & 6 (20.7\%) \\
Qwen3 Max Think. & 19 & 15 (78.9\%) & 4 (21.1\%) \\
Llama 4 Maverick & 28 & 22 (78.6\%) & 6 (21.4\%) \\
\midrule
\textit{Total} & 169 & 130 (76.9\%) & 39 (23.1\%) \\
\bottomrule
\end{tabular}
\end{table}

\begin{table}[htbp]
\centering
\caption{Prompt ablation results across 169 stable divergence events.}
\label{tab:prompt_variance}
\begin{tabular}{lccc}
\toprule
\textbf{Variant} & \textbf{Retained Divergence} & \textbf{Became Unstable} & \textbf{Reversed to Alignment} \\
\midrule
No Role & 124/169 (73.4\%) & 36/169 (21.3\%) & 9/169 (5.3\%) \\
No Research Goal & 113/169 (66.9\%) & 44/169 (26.0\%) & 12/169 (7.1\%) \\
Reframed Task & 136/169 (80.5\%) & 30/169 (17.8\%) & 3/169 (1.8\%) \\
\bottomrule
\end{tabular}
\end{table}

\begin{table}[htbp]
\centering
\caption{Per-model prompt ablation results. Retained = stable divergence preserved; Unstable = choice became position-dependent; Reversed = flipped to author alignment.}
\label{tab:prompt_variance_per_model}
\small
\begin{tabular}{lccccccccccc}
\toprule
 & & \multicolumn{3}{c}{\textbf{No Role}} & \multicolumn{3}{c}{\textbf{No Goal}} & \multicolumn{3}{c}{\textbf{Reframed Task}} \\
\cmidrule(lr){3-5} \cmidrule(lr){6-8} \cmidrule(lr){9-11}
\textbf{Model} & \textbf{Pairs} & \textbf{Ret.} & \textbf{Unst.} & \textbf{Rev.} & \textbf{Ret.} & \textbf{Unst.} & \textbf{Rev.} & \textbf{Ret.} & \textbf{Unst.} & \textbf{Rev.} \\
\midrule
GPT-5.2 Pro & 24 & 83.3 & 12.5 & 4.2 & 79.2 & 20.8 & 0.0 & 91.7 & 4.2 & 4.2 \\
GPT-OSS-120B & 29 & 82.8 & 10.3 & 6.9 & 75.9 & 17.2 & 6.9 & 89.7 & 10.3 & 0.0 \\
Qwen3 Max Think. & 19 & 78.9 & 10.5 & 10.5 & 73.7 & 26.3 & 0.0 & 89.5 & 5.3 & 5.3 \\
Gemini 3 Flash & 19 & 78.9 & 10.5 & 10.5 & 47.4 & 36.8 & 15.8 & 68.4 & 26.3 & 5.3 \\
Claude Haiku 4.5 & 17 & 70.6 & 29.4 & 0.0 & 64.7 & 35.3 & 0.0 & 76.5 & 23.5 & 0.0 \\
Llama 4 Maverick & 28 & 67.9 & 32.1 & 0.0 & 60.7 & 25.0 & 14.3 & 82.1 & 17.9 & 0.0 \\
Claude Opus 4.5 & 21 & 57.1 & 38.1 & 4.8 & 71.4 & 19.0 & 9.5 & 71.4 & 28.6 & 0.0 \\
Gemini 3 Pro & 12 & 58.3 & 33.3 & 8.3 & 50.0 & 41.7 & 8.3 & 58.3 & 41.7 & 0.0 \\
\midrule
\textit{Weighted Mean} & 169 & 73.4 & 21.3 & 5.3 & 66.9 & 26.0 & 7.1 & 80.5 & 17.8 & 1.8 \\
\bottomrule
\end{tabular}
\end{table}

\begin{table}[htbp]
\centering
\caption{Author-preference prediction check on stable divergence events. Retained divergence means the original stable divergence is preserved under the author-preference-prediction framing; unstable means the event is no longer stable across option orderings; reversed means the event flips to author alignment. Parenthesized values are event counts. Gemini 3 Pro is excluded because it was unavailable through our evaluation route at rerun time.}
\label{tab:authorprefpred_stable_divergence}
\small
\begin{tabular}{lcccc}
\toprule
\textbf{Model} & \textbf{Events} & \textbf{Retained Div.} & \textbf{Unstable} & \textbf{Reversed} \\
\midrule
GPT-5.2 Pro & 24 & 62.5\% (15) & 16.7\% (4) & 20.8\% (5) \\
GPT-OSS-120B & 29 & 82.8\% (24) & 13.8\% (4) & 3.4\% (1) \\
Qwen3 Max Think. & 19 & 84.2\% (16) & 10.5\% (2) & 5.3\% (1) \\
Gemini 3 Flash & 19 & 68.4\% (13) & 26.3\% (5) & 5.3\% (1) \\
Claude Haiku 4.5 & 17 & 70.6\% (12) & 29.4\% (5) & 0.0\% (0) \\
Llama 4 Maverick & 28 & 71.4\% (20) & 28.6\% (8) & 0.0\% (0) \\
Claude Opus 4.5 & 21 & 66.7\% (14) & 19.0\% (4) & 14.3\% (3) \\
\midrule
\textit{Weighted Mean} & 157 & 72.6\% (114) & 20.4\% (32) & 7.0\% (11) \\
\bottomrule
\end{tabular}
\end{table}

\begin{table}[htbp]
\centering
\caption{Model-specific retained stable-divergence rates across prompt and author-preference conditions. Wilson 95\% confidence intervals are computed over each model's canonical stable-divergence events. Gemini 3 Pro is not reported for Author-pref because it was unavailable through our evaluation route at rerun time.}
\label{tab:model_retention_ci}
\small
\begin{tabular}{llccc}
\toprule
\textbf{Model} & \textbf{Condition} & \textbf{Retained/Events} & \textbf{Retention} & \textbf{Wilson 95\% CI (\%)} \\
\midrule
\multirow{3}{*}{Gemini 3 Pro}
 & No Role & 7/12 & 58.3\% & [32.0, 80.7] \\
 & No Goal & 6/12 & 50.0\% & [25.4, 74.6] \\
 & Reframed & 7/12 & 58.3\% & [32.0, 80.7] \\
\midrule
\multirow{4}{*}{Gemini 3 Flash}
 & No Role & 15/19 & 78.9\% & [56.7, 91.5] \\
 & No Goal & 9/19 & 47.4\% & [27.3, 68.3] \\
 & Reframed & 13/19 & 68.4\% & [46.0, 84.6] \\
 & Author-pref & 13/19 & 68.4\% & [46.0, 84.6] \\
\midrule
\multirow{4}{*}{Claude Opus 4.5}
 & No Role & 12/21 & 57.1\% & [36.5, 75.5] \\
 & No Goal & 15/21 & 71.4\% & [50.0, 86.2] \\
 & Reframed & 15/21 & 71.4\% & [50.0, 86.2] \\
 & Author-pref & 14/21 & 66.7\% & [45.4, 82.8] \\
\midrule
\multirow{4}{*}{Claude Haiku 4.5}
 & No Role & 12/17 & 70.6\% & [46.9, 86.7] \\
 & No Goal & 11/17 & 64.7\% & [41.3, 82.7] \\
 & Reframed & 13/17 & 76.5\% & [52.7, 90.4] \\
 & Author-pref & 12/17 & 70.6\% & [46.9, 86.7] \\
\midrule
\multirow{4}{*}{GPT-5.2 Pro}
 & No Role & 20/24 & 83.3\% & [64.1, 93.3] \\
 & No Goal & 19/24 & 79.2\% & [59.5, 90.8] \\
 & Reframed & 22/24 & 91.7\% & [74.2, 97.7] \\
 & Author-pref & 15/24 & 62.5\% & [42.7, 78.8] \\
\midrule
\multirow{4}{*}{GPT-OSS-120B}
 & No Role & 24/29 & 82.8\% & [65.5, 92.4] \\
 & No Goal & 22/29 & 75.9\% & [57.9, 87.8] \\
 & Reframed & 26/29 & 89.7\% & [73.6, 96.4] \\
 & Author-pref & 24/29 & 82.8\% & [65.5, 92.4] \\
\midrule
\multirow{4}{*}{Qwen3 Max Think.}
 & No Role & 15/19 & 78.9\% & [56.7, 91.5] \\
 & No Goal & 14/19 & 73.7\% & [51.2, 88.2] \\
 & Reframed & 17/19 & 89.5\% & [68.6, 97.1] \\
 & Author-pref & 16/19 & 84.2\% & [62.4, 94.5] \\
\midrule
\multirow{4}{*}{Llama 4 Maverick}
 & No Role & 19/28 & 67.9\% & [49.3, 82.1] \\
 & No Goal & 17/28 & 60.7\% & [42.4, 76.4] \\
 & Reframed & 23/28 & 82.1\% & [64.4, 92.1] \\
 & Author-pref & 20/28 & 71.4\% & [52.9, 84.7] \\
\bottomrule
\end{tabular}
\end{table}

\begin{table}[htbp]
\centering
\caption{Inter-Model Agreement across three states (alignment, divergent, unstable) and all 80 pairs. Mean agreement: 61.9\% ($\sigma$=6.0\%), mean shared divergence: 15.6\% ($\sigma$=4.4\%). Sorted by agreement rate within intra-family and inter-family groups.}
\label{tab:intermodel_agreement}
\begin{tabular}{lccc}
\toprule
\textbf{Model Pair} & \textbf{Agreement} & \textbf{Shared Div.} & \textbf{Shared Unstable Choice} \\
\midrule
\multicolumn{4}{l}{\textit{Intra-Family}} \\
GPT-5.2 Pro -- GPT-OSS-120B & 80.0\% & 26.2\% & 6.2\% \\
Gemini 3 Pro -- Gemini 3 Flash & 67.5\% & 11.2\% & 3.8\% \\
Claude Opus 4.5 -- Claude Haiku 4.5 & 65.0\% & 12.5\% & 8.8\% \\
\midrule
\multicolumn{4}{l}{\textit{Inter-Family}} \\
Llama 4 Maverick -- GPT-OSS-120B & 71.2\% & 25.0\% & 8.8\% \\
Llama 4 Maverick -- Claude Opus 4.5 & 67.5\% & 21.2\% & 8.8\% \\
GPT-5.2 Pro -- Qwen3 Max & 67.5\% & 18.8\% & 5.0\% \\
GPT-5.2 Pro -- Llama 4 Maverick & 65.0\% & 21.2\% & 5.0\% \\
Claude Opus 4.5 -- GPT-OSS-120B & 65.0\% & 20.0\% & 5.0\% \\
GPT-5.2 Pro -- Claude Opus 4.5 & 65.0\% & 17.5\% & 3.8\% \\
Qwen3 Max -- GPT-OSS-120B & 63.8\% & 20.0\% & 5.0\% \\
GPT-5.2 Pro -- Gemini 3 Pro & 63.8\% & 12.5\% & 2.5\% \\
Gemini 3 Flash -- GPT-OSS-120B & 62.5\% & 16.2\% & 2.5\% \\
Qwen3 Max -- Claude Opus 4.5 & 62.5\% & 16.2\% & 7.5\% \\
GPT-5.2 Pro -- Gemini 3 Flash & 62.5\% & 15.0\% & 0.0\% \\
Llama 4 Maverick -- Claude Haiku 4.5 & 62.5\% & 13.8\% & 10.0\% \\
Gemini 3 Flash -- Claude Opus 4.5 & 61.2\% & 13.8\% & 3.8\% \\
Gemini 3 Pro -- Claude Opus 4.5 & 61.2\% & 12.5\% & 2.5\% \\
Qwen3 Max -- Gemini 3 Flash & 61.2\% & 16.2\% & 3.8\% \\
Claude Haiku 4.5 -- GPT-OSS-120B & 61.2\% & 17.5\% & 5.0\% \\
Gemini 3 Pro -- GPT-OSS-120B & 58.8\% & 13.8\% & 1.2\% \\
GPT-5.2 Pro -- Claude Haiku 4.5 & 57.5\% & 12.5\% & 3.8\% \\
Llama 4 Maverick -- Gemini 3 Flash & 56.2\% & 15.0\% & 2.5\% \\
Qwen3 Max -- Claude Haiku 4.5 & 56.2\% & 12.5\% & 10.0\% \\
Llama 4 Maverick -- Gemini 3 Pro & 55.0\% & 12.5\% & 3.8\% \\
Gemini 3 Flash -- Claude Haiku 4.5 & 53.8\% & 8.8\% & 2.5\% \\
Llama 4 Maverick -- Qwen3 Max & 53.8\% & 16.2\% & 6.2\% \\
Gemini 3 Pro -- Qwen3 Max & 53.8\% & 10.0\% & 3.8\% \\
Gemini 3 Pro -- Claude Haiku 4.5 & 51.2\% & 8.8\% & 2.5\% \\
\bottomrule
\end{tabular}
\end{table}

\begin{table}[htbp]
\centering
\caption{Per-pair human response outcomes for the pilot human baseline. Judgements include fixed choices and abstentions. Author-choice share is computed over fixed choices only, excluding abstentions. Majority denotes whether more participants selected the author-recommended direction or the alternative direction. Equal counts are reported as ties.}
\label{tab:human_per_pair}
\small
\begin{tabular}{lrrrrrl}
\toprule
\textbf{Pair ID} & \textbf{Judg.} & \textbf{Author} & \textbf{Alt.} & \textbf{Abst.} & \textbf{Author Share} & \textbf{Majority} \\
\midrule
2512.22322\_fw0\_d0 & 7 & 3 & 3 & 1 & 50.0\% & Tie \\
2512.22322\_fw1\_d1 & 4 & 2 & 2 & 0 & 50.0\% & Tie \\
2601.02439\_fw0\_d0 & 6 & 4 & 1 & 1 & 80.0\% & Author \\
2601.05175\_fw0\_d0 & 4 & 3 & 1 & 0 & 75.0\% & Author \\
2601.05175\_fw3\_d1 & 4 & 4 & 0 & 0 & 100.0\% & Author \\
2601.09680\_fw3\_d0 & 6 & 2 & 3 & 1 & 40.0\% & Alternative \\
2601.13232\_fw0\_d1 & 5 & 3 & 1 & 1 & 75.0\% & Author \\
2601.14525\_fw0\_d1 & 4 & 3 & 0 & 1 & 100.0\% & Author \\
2601.14525\_fw5\_d0 & 5 & 3 & 1 & 1 & 75.0\% & Author \\
2601.16175\_fw0\_d0 & 6 & 2 & 3 & 1 & 40.0\% & Alternative \\
2601.18207\_fw4\_d0 & 5 & 4 & 0 & 1 & 100.0\% & Author \\
2601.18207\_fw8\_d1 & 5 & 0 & 5 & 0 & 0.0\% & Alternative \\
2601.21468\_fw0\_d1 & 7 & 6 & 1 & 0 & 85.7\% & Author \\
2601.21468\_fw1\_d1 & 8 & 3 & 5 & 0 & 37.5\% & Alternative \\
2601.22930\_fw0\_d1 & 4 & 4 & 0 & 0 & 100.0\% & Author \\
2602.02366\_fw0\_d0 & 3 & 0 & 3 & 0 & 0.0\% & Alternative \\
2602.02366\_fw2\_d1 & 4 & 4 & 0 & 0 & 100.0\% & Author \\
2602.03955\_fw2\_d0 & 3 & 1 & 0 & 2 & 100.0\% & Author \\
2602.07491\_fw3\_d1 & 4 & 4 & 0 & 0 & 100.0\% & Author \\
2602.07755\_fw0\_d1 & 4 & 1 & 3 & 0 & 25.0\% & Alternative \\
2602.09163\_fw1\_d1 & 3 & 0 & 2 & 1 & 0.0\% & Alternative \\
2602.09163\_fw3\_d1 & 5 & 1 & 2 & 2 & 33.3\% & Alternative \\
2602.10067\_fw0\_d1 & 2 & 1 & 1 & 0 & 50.0\% & Tie \\
\bottomrule
\end{tabular}
\end{table}

\begin{table}[htbp]
\centering
\caption{Aggregate human-baseline results for the full pilot set and two stricter query subsets. Fixed responses are non-abstaining judgements. Pooled choice rates weight each fixed human response equally, while per-query macro-average choice rates first compute author and alternative shares within each pair and then average across pairs. Pair majorities are computed from fixed responses only.}
\label{tab:human_aggregate}
\small
\setlength{\tabcolsep}{3pt}
\begin{tabular}{lrrrrccccccc}
\toprule
& & & & & \multicolumn{2}{c}{\textbf{Pooled choice}} & \multicolumn{2}{c}{\textbf{Macro choice}} & \multicolumn{3}{c}{\textbf{Pair majority}} \\
\cmidrule(lr){6-7} \cmidrule(lr){8-9} \cmidrule(lr){10-12}
\textbf{Subset} & \textbf{Pairs} & \textbf{Judg.} & \textbf{Fixed} & \textbf{Abst.} & \textbf{Author} & \textbf{Alt.} & \textbf{Author} & \textbf{Alt.} & \textbf{Author} & \textbf{Alt.} & \textbf{Tie} \\
\midrule
All 23 pairs & 23 & 108 & 95 & 13 & 61.1\% & 38.9\% & 61.6\% & 38.4\% & 12 & 8 & 3 \\
$\geq 3$ fixed responses & 20 & 100 & 90 & 10 & 62.2\% & 37.8\% & 63.3\% & 36.7\% & 11 & 7 & 2 \\
No query-quality abstentions & 13 & 56 & 55 & 1 & 58.2\% & 41.8\% & 55.6\% & 44.4\% & 6 & 5 & 2 \\
\bottomrule
\end{tabular}
\end{table}

\begin{table}[htbp]
\centering
\caption{Participant metadata for the pilot human baseline. Participants are anonymized. Fixed responses are non-abstaining judgements. Six participants reported a Bachelor's degree, one reported a Master's degree, one reported a PhD, one reported a postdoctoral role, and one selected Other. Median self-reported ML experience was 3 years, with a range from 0 to 13 years and a mean of 4.5 years.}
\label{tab:human_participants}
\small
\begin{tabular}{llrrrr}
\toprule
\textbf{Participant} & \textbf{Highest degree / role} & \textbf{Years ML exp.} & \textbf{Judg.} & \textbf{Fixed} & \textbf{Abst.} \\
\midrule
P01 & PhD & 2 & 3 & 3 & 0 \\
P02 & Bachelor's & 3 & 10 & 10 & 0 \\
P03 & Bachelor's & 6 & 23 & 16 & 7 \\
P04 & Master's & 4 & 10 & 10 & 0 \\
P05 & Bachelor's & 13 & 5 & 3 & 2 \\
P06 & Other & 1 & 19 & 18 & 1 \\
P07 & Bachelor's & 0 & 5 & 4 & 1 \\
P08 & Bachelor's & 3 & 23 & 21 & 2 \\
P09 & Postdoc & 11 & 5 & 5 & 0 \\
P10 & Bachelor's & 2 & 5 & 5 & 0 \\
\bottomrule
\end{tabular}
\end{table}

\begin{table}[htbp]
\centering
\caption{Abstention reasons in the pilot human baseline. Shares use all 108 submitted human judgements as the denominator.}
\label{tab:human_abstention_reasons}
\small
\begin{tabular}{lrr}
\toprule
\textbf{Abstention reason} & \textbf{Count} & \textbf{Share of all judgements} \\
\midrule
Options do not make sense & 6 & 5.6\% \\
Context unclear / insufficient & 4 & 3.7\% \\
Outside domain expertise & 2 & 1.9\% \\
Other & 1 & 0.9\% \\
\midrule
\textit{Total} & 13 & 12.0\% \\
\bottomrule
\end{tabular}
\end{table}

\begin{table}[htbp]
\centering
\caption{Five-category taxonomy for research direction labelling on both ground truth and distractor directions in the 45 stable divergence pairs. When a future work introduces a stronger model capability that is external to the paper's main contribution, it is coded as Scaling/Generalization/Robustness rather than Methodological Innovation.}
\label{tab:taxonomy}
\begin{tabular}{lp{8cm}}
\toprule
\textbf{Category} & \textbf{Description} \\
\midrule
Infrastructure, Data \& Benchmarking & Build datasets, benchmarks, tooling, or evaluation infrastructure. \\
Mechanistic / Interpretability / Theory & Explain why a result holds, probe internals, or formalize the phenomenon. \\
Methodological \& Architectural Innovation & Modify the algorithm, training objective, architecture, or system design. \\
Scaling, Generalization \& Robustness & Test the existing method across scales, domains, settings, or stress conditions. \\
Application \& Downstream Integration & Apply the method to a concrete downstream workflow, real-world pipeline, domain, or agentic/use-case setting. \\
\bottomrule
\end{tabular}
\end{table}

\begin{table}[htbp]
\centering
\caption{GT-to-distractor category transitions across 169 stable divergence events. The \% of Total reflects each transition's share of all 169 events and the \% within GT reflects its share within the corresponding ground-truth category's divergence events.}
\label{tab:transitions}
\begin{tabular}{llccc}
\toprule
\textbf{GT Category} & \textbf{Distractor Category} & \textbf{Count} & \textbf{\% of Total} & \textbf{\% within GT} \\
\midrule
\multirow{5}{*}{Methodological (53)}
 & Infrastructure & 25 & 14.8\% & 47.2\% \\
 & Scaling & 23 & 13.6\% & 43.4\% \\
 & Methodological & 2 & 1.2\% & 3.8\% \\
 & Mechanistic & 2 & 1.2\% & 3.8\% \\
 & Application & 1 & 0.6\% & 1.9\% \\
\midrule
\multirow{4}{*}{Infrastructure (75)}
 & Methodological & 30 & 17.8\% & 40.0\% \\
 & Infrastructure & 26 & 15.4\% & 34.7\% \\
 & Scaling & 10 & 5.9\% & 13.3\% \\
 & Mechanistic & 9 & 5.3\% & 12.0\% \\
\midrule
\multirow{3}{*}{Application (25)}
 & Scaling & 12 & 7.1\% & 48.0\% \\
 & Application & 8 & 4.7\% & 32.0\% \\
 & Methodological & 5 & 3.0\% & 20.0\% \\
\midrule
\multirow{3}{*}{Scaling (14)}
 & Mechanistic & 5 & 3.0\% & 35.7\% \\
 & Scaling & 5 & 3.0\% & 35.7\% \\
 & Infrastructure & 4 & 2.4\% & 28.6\% \\
\midrule
Mechanistic (2)
 & Scaling & 2 & 1.2\% & 100.0\% \\
\bottomrule
\end{tabular}
\end{table}

\begin{table}[htbp]
\centering
\caption{Selected dominant stable-divergence transitions in the multi-model subset. This table restricts to the 144 stable-divergence events from pairs where at least three models stably diverged, and reports the most frequent distractor categories for the largest author-recommended direction categories. The same broad transition patterns observed in the full 169-event table remain visible under this multi-model filter.}
\label{tab:transitions_selected}
\begin{tabular}{llcc}
\toprule
\textbf{GT Category} & \textbf{Distractor Category} & \textbf{Events} & \textbf{\% within GT} \\
\midrule
\multirow{2}{*}{Methodological (32)}
 & Infrastructure & 18 & 56.2\% \\
 & Scaling & 14 & 43.8\% \\
\midrule
\multirow{2}{*}{Infrastructure (74)}
 & Methodological & 30 & 40.5\% \\
 & Infrastructure & 25 & 33.8\% \\
\midrule
\multirow{2}{*}{Application (25)}
 & Scaling & 12 & 48.0\% \\
 & Application & 8 & 32.0\% \\
\bottomrule
\end{tabular}
\end{table}

\begin{table}[htbp]
\centering
\caption{MCQ pilot performance on the future-work selection task (n=35 queries, 3 samples each). We evaluate nine models across four provider families and report majority-vote accuracy, format compliance, T=0 determinism, and A/B/C choice distribution compared to the ground-truth distribution. Author alignment in this setting ranged from 34.3\% for Claude Sonnet 4.5 to 60.0\% for Gemini 3 Pro. Choice distributions reveal position sensitivity. While ground truth is concentrated at position A, several models over-select position C.}
\label{tab:mcq_main_results}
\begin{tabular}{llcccc}
\toprule
\textbf{Family} & \textbf{Model} & \textbf{Accuracy} & \textbf{Format Compl.} & \textbf{T=0 Det.} & \textbf{Choice Dist.} \\
\midrule
\multirow{3}{*}{Anthropic}
 & Claude Opus 4.5   & 54.3\% & 100\% & 74.3\% & 23/36/46 \\
 & Claude Sonnet 4.5 & 34.3\% & 100\% & 54.3\% & 15/30/60 \\
 & Claude Haiku 4.5  & 45.7\% & 100\% & 85.7\% & 20/41/44 \\
\midrule
\multirow{2}{*}{Google}
 & Gemini 3 Pro   & 60\% & 99\% & 91.4\% & 41/30/34 \\
 & Gemini 3 Flash & 57.1\% & 100\% & 74.3\% & 38/35/32 \\
\midrule
\multirow{2}{*}{OpenAI}
& GPT-5.2 Pro  & 40\% & 100\% & 75\% & 22/21/62 \\
& gpt-oss-120b & 51.4\% & 100\% & 71.4\% & 32/21/52 \\
\midrule
\multirow{2}{*}{Open Source}
& Llama 4 Maverick & 45.7\% & 100\% & 45.7\% & 34/32/39 \\
& Qwen3 Max Think. & 42.9\% & 100\% & 85.7\% & 23/17/65 \\
\midrule
\multicolumn{2}{l}{\textit{Ground Truth Count}} & & -- & -- & 57/30/18 \\
\bottomrule
\end{tabular}
\end{table}

\begin{table}[htbp]
    \centering
    \caption{MCQ accuracy by ground-truth position for the position-control audit (n=35 queries per position). To disentangle content-level evaluation from position heuristics, we evaluated three models on variants where each query appears with the ground-truth option at A, B, and C. Accuracy varies by ground-truth position, especially for Claude Opus 4.5 and Qwen3 Max Think., motivating our adoption of pairwise comparison with balanced position control.}
    \label{tab:mcq_position_bias}
    \begin{tabular}{lcccc}
    \toprule
    \textbf{Model} & \textbf{GT at A} & \textbf{GT at B} & \textbf{GT at C} & \textbf{Range(\%)}\\
    \midrule
    Gemini 3 Flash    & 65.7\% & 60.0\% & 68.6\% & 8.6 \\
    Claude Opus 4.5   & 42.9\% & 60.0\% & 71.4\% & 28.5 \\
    Qwen3 Max Think.  & 25.7\% & 51.4\% & 68.6\% & 42.9 \\
    \bottomrule
    \end{tabular}
\end{table}

\end{document}